# PREDICTING CLASS-IMBALANCED BUSINESS RISK USING RESAMPLING, REGULARIZATION, AND MODEL EMSEMBLING ALGORITHMS

Yan Wang[1], Xuelei Sherry Ni[2]

[1]Graduate College, Kennesaw State University, Kennesaw, USA
[2]Department of Statistics and Analytical Sciences, Kennesaw State University, Kennesaw, USA

*ABSTRACT*

*We aim at developing and improving the imbalanced business risk modeling via jointly using proper evaluation criteria, resampling, cross-validation, classifier regularization, and ensembling techniques. Area Under the Receiver Operating Characteristic Curve (AUC of ROC) is used for model comparison based on 10-fold cross validation. Two undersampling strategies including random undersampling (RUS) and cluster centroid undersampling (CCUS), as well as two oversampling methods including random oversampling (ROS) and Synthetic Minority Oversampling Technique (SMOTE), are applied. Three highly interpretable classifiers, including logistic regression without regularization (LR), L1-regularized LR (L1LR), and decision tree (DT) are implemented. Two ensembling techniques, including Bagging and Boosting, are applied on the DT classifier for further model improvement. The results show that, Boosting on DT by using the oversampled data containing 50% positives via SMOTE is the optimal model and it can achieve AUC, recall, and F1 score valued 0.8633, 0.9260, and 0.8907, respectively.*

## 1. INTRODUCTION

Risk modeling can discriminate the risky business from the non-risky companies, thus can guide the financial institutions to make decisions when processing loan or credit applications [1] [2]. Logistic regression (LR) is a frequently used technique for risk classifications since it is conceptually simple and explainable and has been demonstrated to be powerful in many studies [3] [4] [5]. Decision tree (DT) is also widely used as it has strong interpretability and relatively straightforward structures compared to complicated models such as neural networks [6].

However, in real-life applications of risk models, many data are imbalanced. That is, the distributions in each class are not uniform [7]. Imbalanced data result in several problems and challenges to existing algorithms that have been shown to be effective on the balanced data [8] [9] [10]. The problems arising from modeling imbalanced data mainly contain the following three categories: (1) the usage of improper model evaluation metric; (2) data rarity due to the lack of observations in the rare class; and (3) the usage of weak classifiers or classifiers without regularization [11] [12] [13]. To handle the problems mentioned above, researchers have proposed several effective methodogies, including using more appropriate evaluate metrics, resampling (oversampling or undersampling) with different ratios between positives and negatives, using cross validation in the right way, classifier regularization such as L1-regularized Logistic Regression (L1LR) and model ensembling [14] [15] [16] [17].

Motivated by previous research, we design a modeling work flow in this study, aiming at developing a good model for imbalanced risk classifications by using the aforementioned class-imbalance-handling methods simultaneously. The utilization of the above-mentioned algorithms jointly is a major strength of this study, since there are few previous research focusing on using these techniques in one modeling design. The details of the modeling flow are described in Figure 3.

This paper is structured as follows. The relevant techniques used for handling class imbalance are reviewed in Section 2. Section 3 describes the experimental design. Experimental results are discussed in Section 4. Section 5 addresses the conclusions.

## 2. METHODOLOGIES FOR HANDLING CLASS IMBALANCE

In this section, the methodologies used in this study for classifying imbalanced targets are reviewed.

### 2.1. Selection of the proper evaluation metrics

Classification accuracy is generally considered the optimal measure for binary classification. However, it is no longer a good evaluation metric for models built on imbalanced data because accuracy will bias towards the majority class. Instead, Area Under the Receiver Operating Characteristic Curve (AUC of ROC) is more appropriate since it does not place more emphasis on one class over another and it is not biased against the minority class [7] [18] [19]. Moreover, ROC curve is independent of the positive-negative ratio in target, making AUC a suitable measure for comparing different classifiers when the positive-negative ratio varies [16]. Let P and N represent the total number of positive cases (i.e., risky business) and negatives cases (i.e., non-risky business), respectively. Let True Positive (TP) and False Positive (FP) denote those identified as risky business correctly or wrongly, respectively. Similarly, we denote True Negative (TN) and False Negative (FN) as those identified as non-risky business correctly or wrongly, respectively. Then recall (shown in Equation 1) measures the fraction of risky business correctly classified and precision (described in Equation 2) measures the fraction of objects classified as risky business that are truly risky. Considering that a FN error may signify the loss, recall is used as the secondary evaluation criterion in this study and precision is weighted less. Besides, F1 score (described in Equation 3) is utilized as the third evaluation metric as it represents the harmonic mean of precision and recall [20].

$$\text{Recall} = \frac{TP}{TP + FN} \qquad (1)$$

$$\text{Precision} = \frac{TP}{TP + FP} \qquad (2)$$

$$\text{F1 score} = \frac{2 * precision * recall}{precision + recall} \qquad (3)$$

## 2.2. Resampling methods

Resampling, aiming at eliminating or minimizing the rarity by altering the distribution of training samples, is a widely used while effective technique in the rare event prediction [21] [22]. The resampling technique mainly includes two approaches: undersampling and oversampling. The undersampling means removing observations in the majority class while in oversampling, we duplicate the records in the minority class [18]. Figure 1 illustrates the graphical representations of the above-mentioned two approaches. Both could balance the class distribution, thus make the binary classifiers more robust. Although undersampling is often used when the size of data is sufficient while oversampling is preferred when the data is relatively small, there is no absolute advantage of one resampling method over another. Application of oversampling or undersampling algorithms depends on different research cases and dataset used [23].

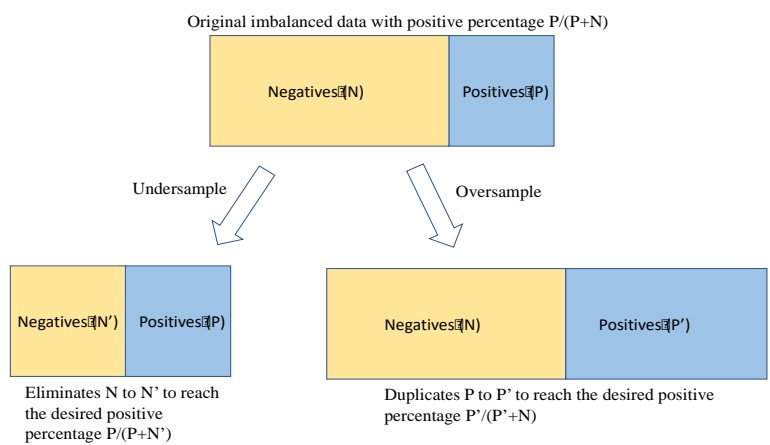

Figure 1. Graphical representation of the undersampling and oversampling algorithms

In this paper, two undersampling techniques -- random undersampling (RUS) and cluster centroid undersampling (CCUS) -- are implemented. RUS eliminates the examples in the majority class (usually defined as the negatives in risk modeling) randomly and uniformly while keeping all the observations from the minority class (usually defined as the positives in risk modeling) to reach a desired positive percentage [23]. In CCUS, all observations in the minority class of the original data are kept. Then, the clusters of the majority class are found using $K$-means algorithms and only the $K$ cluster centroids are kept for replacing the points from the majority class [24]. The $K$ in CCUS is set to the value that can reach the desired percentage of positives.

In addition, two oversampling methods -- random oversampling (ROS) and Synthetic Minority Oversampling Technique (SMOTE) -- are applied. Based on the similar idea as in RUS, in ROS, the observations in the minority class are duplicated randomly and uniformly while all the examples from the majority class are kept to reach the desired positive ratio [25]. On the other hand, SMOTE does not duplicate the records in the minority class. Instead, SMOTE could add newly synthesized instances (belonging to the minority class) to the original data. And these newly generated examples should be similar to the positives from the original data with respect to Euclidean distance. Figure 2 shows the details of the SMOTE algorithm. In subplot (1) of Figure 2, the originally imbalanced data is represented with a majority of negatives (blue points) and a minority of positives (orange points). In subplot (2), we randomly select a positive instance (green point) and then find its $k$-nearest neighbours among all the positives. In our study, we set *k* to be five and these neighbours are denoted by the block dots marked with numbers. Next, in subplot (3), among the five neighbours, one neighbour is randomly selected. The figure shows the illustrative case when neighbour numbered 3 is selected. Then a new data point (denoted by the red dot) is generated along the straight line connecting the neighbour numbered 3 and the green point. Finally, this newly synthesized instance is labelled as positive and added to the original dataset [26]. The procedure repeated for the positives to get the data with the desired positive percentage.

## 2.3. Resampling to reach different positive percentages

During the resampling algorithm, one hyper-parameter we need to tune is the ratio between positives and negatives (or, equivalently, the positive percentage in the training data). It is shown by previous research that the best ratio (or, the best positive percentage) heavily depends on the data and the models that are used, making it worth trying different percentages and selecting the optimal one rather than training models using the same positive percentage [27]. In this study, we use the resampling methods to obtain a series of training datasets containing the positive percentages ranging from 10% to 90% with a step of 10%.

## 2.4. Using cross-validation in the proper way

Similar as comparing performance of classifiers on the balanced data, *k*-fold cross validation is also frequently used when evaluating performance of models on class-imbalanced data. The most important thing worthy to be emphasized is that *k*-fold cross validation should always be done before, rather than after performing resampling strategies on the training data (see details in Figure 3). If *k*-fold cross validation is applied after resampling, it will cause the overlap of the training sets as well as leading to the over-fitting issue. On the contrary, if we perform *k*-fold cross-validation before resampling, randomness can be introduced into the dataset and the over-fitting problem can be reduced [28]. In our study, we set the value of *k* to be 10 since many studies have shown that 10-fold cross validation can obtain efficiently computational time while good model performance.

## 2.5. Regularization on logistic regression

It is shown that classic LR is not so robust without weighting and regularization when being used on rare event data [29]. L1-regularized LR (L1LR), which is one of the widely used regularized versions of LR, has been shown to outperform the classic LR in modeling imbalanced data and high-dimensional data [1] [17] [30] [31].

L1LR works as follows. Suppose we have a training set with $X \in R^{n \times d}$, where *n* is the number of observations and *d* denotes the number of features. For every instance $x_i \in R^d$, the dependent variable $y_i$ follows a *Bernoulli* distribution with probability of $p_i$ being 1 and probability of (1-$p_i$) being 0. LR models the relationship between each instance $x_i$ with its expected outcome $p_i$ using Equation 4, where $(\beta_0, \beta_1, ..., \beta_n)$ is the parameter vector and can be denoted by a vector $\boldsymbol{\beta}$. By adding a regularization (penalty) term $\lambda ||\boldsymbol{\beta}||_1$ into the objective function (i.e., the log-likelihood function) of LR, we get the optimization problem for L1LR defined in Equation 5, where $\lambda$ is a hyper-parameter to control the regularization and $||\boldsymbol{\beta}||_1$ is the $L_1$ norm of the parameters.

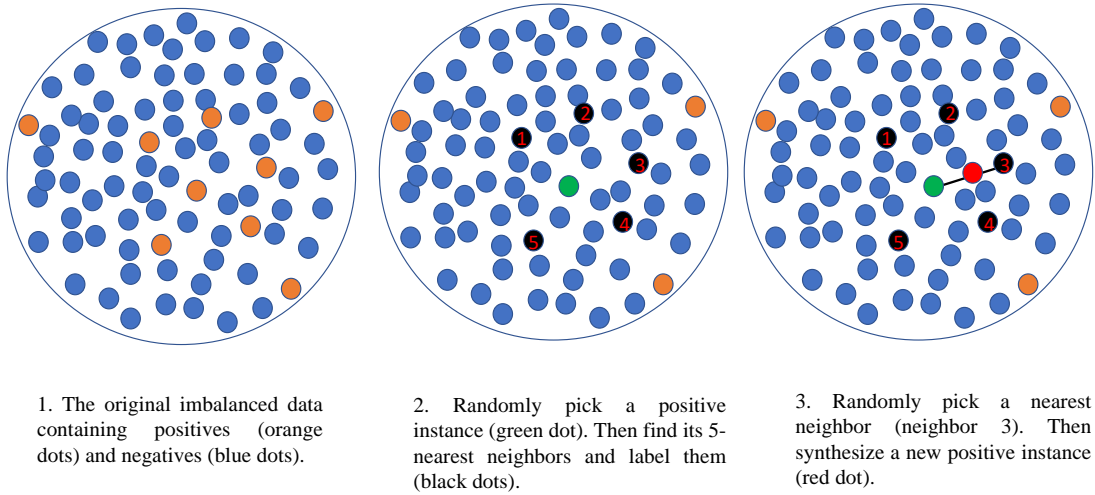

1. The original imbalanced data containing positives (orange dots) and negatives (blue dots).

2. Randomly pick a positive instance (green dot). Then find its 5-nearest neighbors and label them (black dots).

3. Randomly pick a nearest neighbor (neighbor 3). Then synthesize a new positive instance (red dot).

Figure 2. Graphical representation of the SMOTE algorithm

$$p_i = \frac{\exp(x_i\beta)}{1 + \exp(x_i\beta)} \qquad (4)$$

$$min \sum_{i=1}^{n} -\log p_i \, l(y_i|x_i; \beta) + \lambda ||\beta||_1 \qquad (5)$$

## 2.6. Decision tree and model ensembling

Some studies point out that one of the reasons DT is not efficient for imbalanced data is, in the building of DT, the instance space is partitioned into smaller and smaller spaces, making it difficult to find the regularities [16]. However, there are still several studies showing the promising utilization and the robustness of DT when being used on the class-imbalanced data [32] [33]. Considering that DT is widely accepted in the financial domain because of its strong interpretability as well as its robustness in the above-mentioned studies, in our experiment we use it as one of the classifiers, aiming to investigate whether it is still powerful in class-imbalanced risk classification. For the DT algorithm, it searches the optimal splits on input variables based on different criteria such as entropy or Gini index. Gini index is used in this paper and its calculation for a given node is defined in Equation 6, where $C$ is the number of classes in the dependent variable and p(code) is the relative frequency of class c at the node [34].

$$Gini(node) = 1 - \sum_{c=1}^{C} [p(c|node)]^2 \qquad (6)$$

Many studies have confirmed that comparing to a single learning algorithm, ensemble methods can obtain better performance in rare event predictions by using multiple base learners [35] [36]. Two widely used ensembling approaches are Bagging and Boosting. Bagging builds a series of parallel while independent base classifiers using bootstrap subsets of the training set. On the contrary, Boosting builds a series of base classifiers sequentially and dependently. As a result, the subset used for building the current base classifier is not generated by bootstrap sampling. Instead, the algorithm changes the instance weights for the current classifier based on the performance of the previous classifiers. Instances that are misclassified by previous classifiers would be weighted more in the current classifier [37] [38]. In the end, both methods use a majority voting logic for the final prediction. In this study, we consider both Bagging and Boosting approaches based on the DT classifier, aiming to investigate whether we can obtain a better risk model through ensembling. It is worth noting that before performing model ensembling, we focus on getting a DT model as accurate as possible by identifying the proper resampling method along with a properly resampled positive ratio, as described in Sections 2.2 and 2.3. During the processing of Bagging and Boosting on DT classifier, Gini index is also used to look for candidate variables for splitting each node in each DT model. Moreover, Gini reduction is used to rank the variable importance for business risk after the optimal model is identified.

## 3. EXPERIMENTAL DESIGN

### 3.1. Data description and pre-processing

The dataset provided by the US national Credit Bureau is used in our study to develop and evaluate the risk models. The data contains commercial information of over 10 million de-identified U.S. companies from 2006 to 2014. Example commercial information includes business location, business size, liens, industry account activities, and liabilities. The bankruptcy indicator represents the status of the business: 0 denotes the business is still running while 1 means the business went to bankruptcy. To evaluate the risk of going bankruptcy of the companies in 2014, a new target variable *RiskInd* is calculated based on the change of the bankruptcy indicator. In other words, in our study, the positives (i.e., risky business) are businesses which have changed the value of bankruptcy indicator from 0 in 2013 to 1 in 2014 while the negatives (i.e., non-risky business) are those having bankruptcy indicator valued 0 in both 2013 and 2014. Observations that have bankruptcy indicator valued 1 in both 2013 and 2014 are excluded in our modeling.

A series of data pre-processing procedures were applied sequentially as follows: (1) We performed 10-fold cross validation in our study. In each of the 10-fold cross validation, 90% data (i.e., 18,000 observations) was used as the training set while 10% (i.e., 2,000 records) was used as the validation set; (2) Variables with more than 70% missing were removed; (3) For both training and validation sets, missing values were imputed by the median of the variable from the training set; (4) For both training and validation sets, variables were standardized by using the means and the standard deviations of the variable from the training set. The percentage of positives is about 7.4% and 36 independent variables are kept for further analysis.

### 3.2. Methodology

Figure 3 shows the workflow of our numerical experiments by using oversampling as the illustrative example. There are mainly five stages in the workflow. In stage 1, the experiment starts by investigating the effect of different resampling methods on different classifiers. As described in Section 2.3, in the oversampling experiments, ROS and SMOTE are implemented on the training data to reach a series of resampled sets with different positive percentages: 10% to 90% with a step of 10%. The same procedure is applied in the undersampling experiments, except that RUS and CCUS are implemented. In stage 2, three classifiers including LR, L1LR, and DT are applied on each of the resampled training sets. Then in stage 3, the model performance is evaluated and compared on the validation set using AUC, recall, and F1 score. Stages 1, 2, and 3 are repeated 10 times in 10-fold cross validation and the optimal resampling method as well as the best resampled positive percentage are selected for each of the three classifiers based on the average cross-validated results. Therefore, in this stage, the most accurate base classifier is built. Next, in stage 4, Bagging and Boosting are implemented on DT using the training data with the optimal positive percentage and the best resampling method from stage 3. The purpose is to examine whether the performance of an individual DT can be further improved by model ensembling. Finally, in stage 5, the optimal model is selected and the important features related to risk classifications are identified.

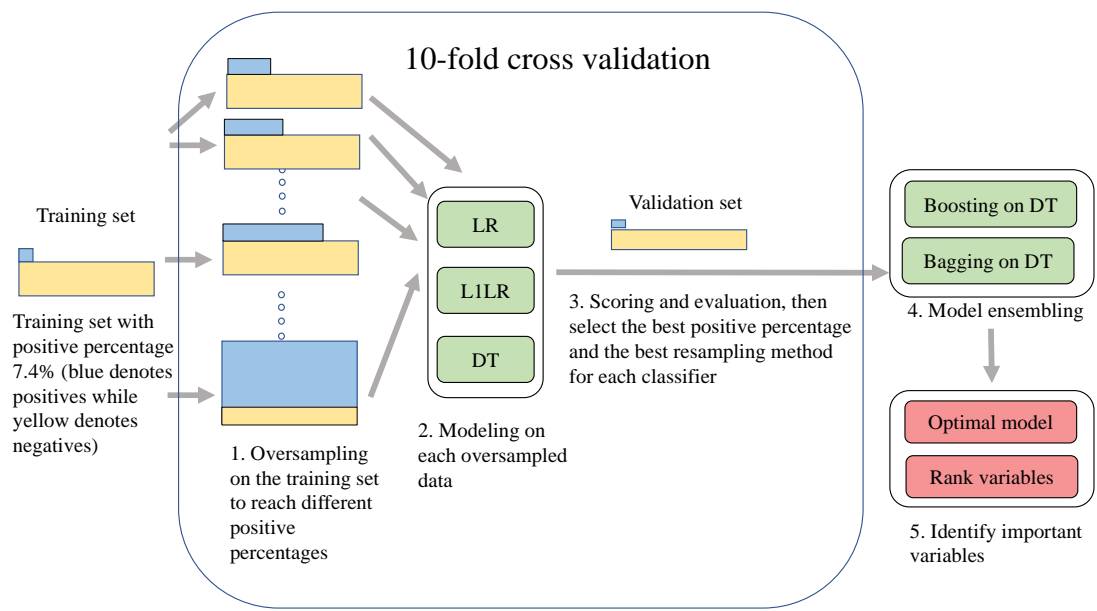

Figure 3. The workflow of the experimental design by using oversampling as the illustrative example

## 4. EXPERIMENTAL RESULTS

With respect to the analysis tools, we use SAS (version 9.4) for the data pre-processing. Python (version 3.5) is used for the rest of our analysis including resampling, 10-fold cross validation, implementation of LR, L1LR, and DT, as well as Bagging and Boosting on DT. Our experiments are implemented using a local desktop configured with a 3.3 GHz Intel Core i7 processor, 16GB RAM, and MacOS system.

We started our experiments by first investigating the effect of different resampling methods on model performance, and the results came out in the end of stage 3, as described in Figure 3. Taking the 50% positive percentage value as an illustrative example, Figure 4 displays the visualization of the resampled data with the first two principal components through principal component analysis (PCA). The figure clearly shows that the sizes of the data obtained by RUS and CCUS are obviously smaller than those obtained by ROS and SMOTE. Although all the four resampled sets have 50% positives, the distributions of the positives and the negatives are different. The main difference is that the data obtained from SMOTE contains many newly generated points while the rest three samples only contain the points from the original data. It is because SMOTE algorithm synthesizes new points as illustrated by Figure 2.

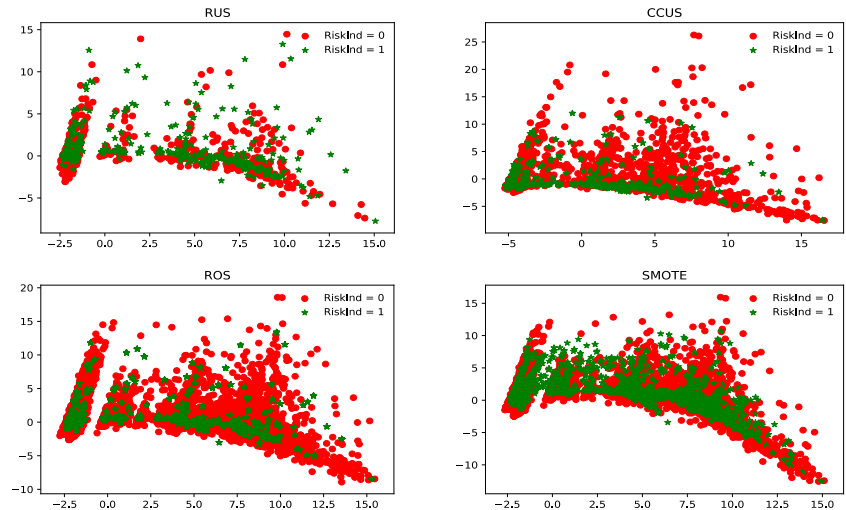

Figure 4. PCA visualization of the training set with 50% positives obtained from four resampling methods: RUS, CCUS, ROS, and SMOTE

In the end of stage 3 in the workflow, the average AUC of each classifier, LR, L1LR, or DT, is calculated from 10-fold cross validation and the results are illustrated by Figures 5, 6, and 7, respectively. Overall, we see that, the different positive percentages result in very different model performance. Furthermore, different resampling methods affect the model performance in a various extent. In general, SMOTE outperforms the rest three resampling methods on all three classifiers across almost all training sets with different positive percentages. The highest AUC is obtained by SMOTE when the training set has 20%, 20%, and 50% risky business on LR, L1LR, and DT, respectively (as marked by a circle in Figures 5, 6, and 7). We label the corresponding models as LR_SMOTE_20%, L1LR_SMOTE_20%, and DT_SMOTE_50%, respectively.

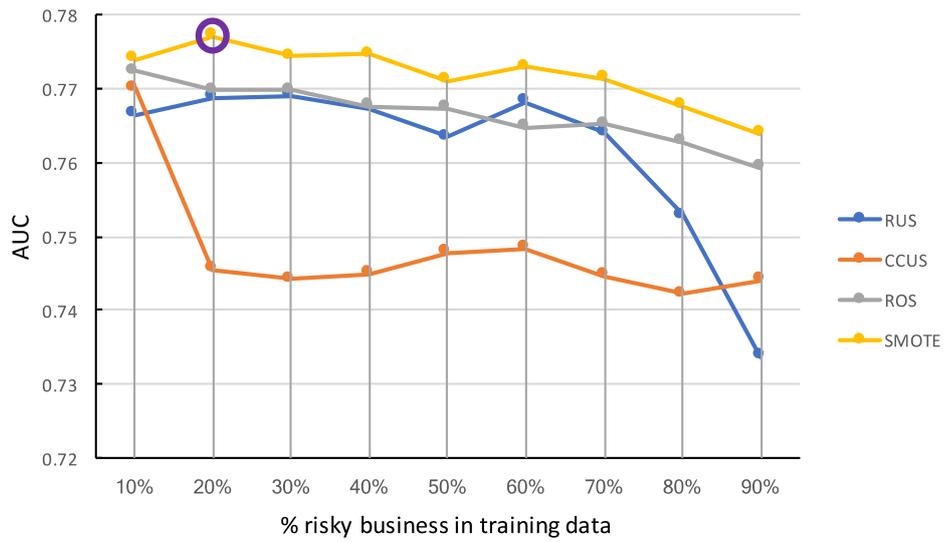

Figure 5. AUC of LR versus various positive rates in the training data across different resampling methods

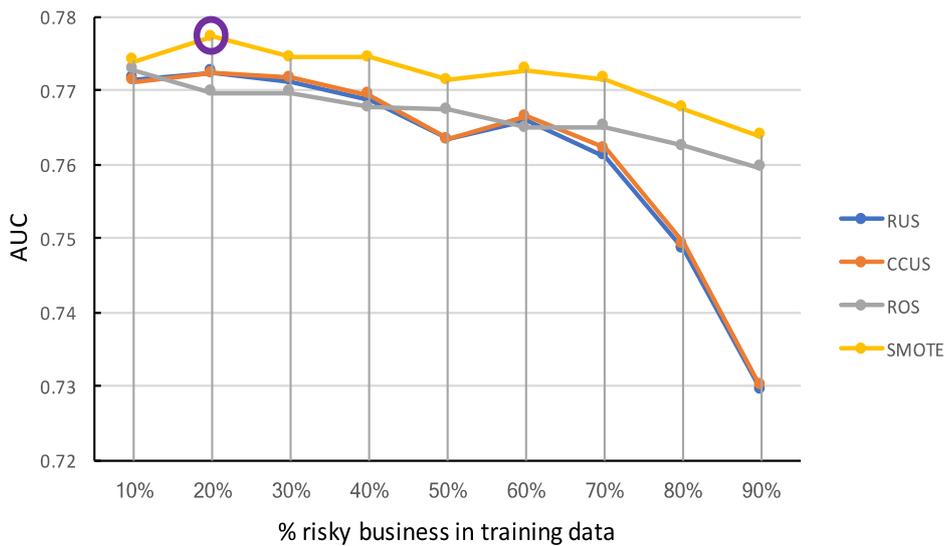

Figure 6. AUC of L1LR versus various positive rates in the training data across different resampling methods

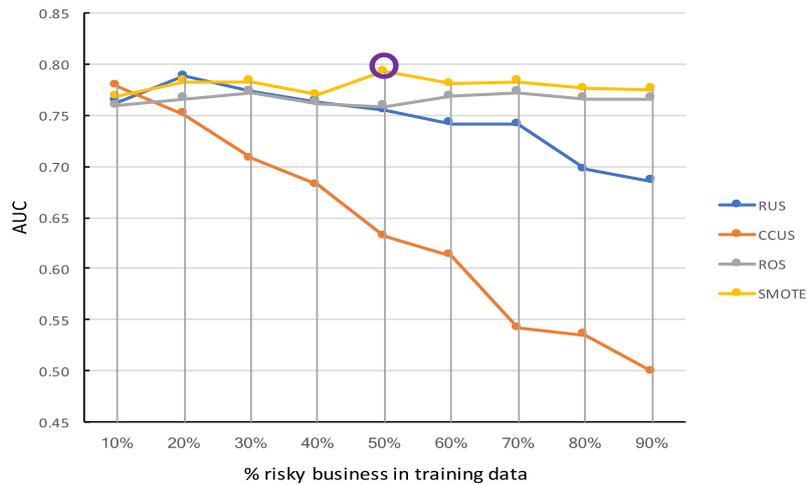

Figure 7. AUC of DT versus various positive rates in the training data across different resampling methods

To investigate the three models including LR_SMOTE_20%, L1LR_SMOTE_20%, and DT_SMOTE_50% in more detail, we compare their performance using more evaluation criteria described in Section 2.1. Furthermore, to identify the benefits of resampling, we have implemented LR, L1LR, and DT on the original data (i.e., percentage of positives is 7.4%) and the resulting models are labelled as LR_original_7.4%, L1LR_original_7.4%, and DT_original_7.4%, respectively. All the above-mentioned results are illustrated in Figure 8. By comparing performance between LR_original_7.4% and LR_SMOTE_20%, between L1LR_original_7.4% and L1LR_SMOTE_20%, and between DT_original_7.4% and DT_SMOTE_50%, we found that resampling strategy can improve AUC, recall, and F1 score in all these three classifiers, with recall having the most obvious increase. Surprisingly, LR and L1LR have very similar performance by using all the evaluation criteria, no matter whether the training data is resampled or not. This indicates that the LR with L1-regularization does not show its superiority over LR without regularization in our study.

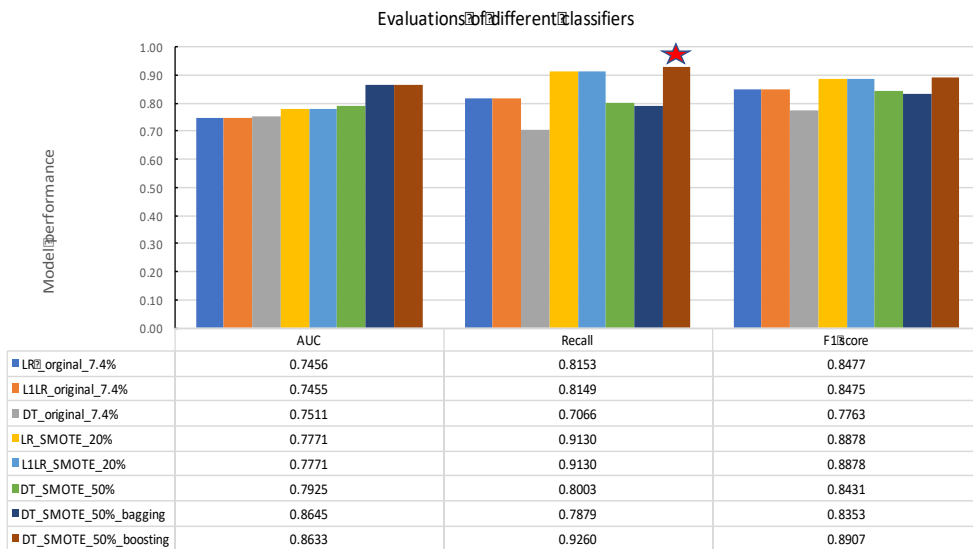

|  | AUC | Recall | F1 score |
|---|---|---|---|
| LR_orginal_7.4% | 0.7456 | 0.8153 | 0.8477 |
| L1LR_original_7.4% | 0.7455 | 0.8149 | 0.8475 |
| DT_original_7.4% | 0.7511 | 0.7066 | 0.7763 |
| LR_SMOTE_20% | 0.7771 | 0.9130 | 0.8878 |
| L1LR_SMOTE_20% | 0.7771 | 0.9130 | 0.8878 |
| DT_SMOTE_50% | 0.7925 | 0.8003 | 0.8431 |
| DT_SMOTE_50%_bagging | 0.8645 | 0.7879 | 0.8353 |
| DT_SMOTE_50%_boosting | 0.8633 | 0.9260 | 0.8907 |

Figure 8. Comparison across different base classifiers and ensemble modelling using AUC, recall, and F1 score

In the stage 4 illustrated by Figure 3, two model ensembling techniques, Bagging and Boosting, are applied on DT_SMOTE_50%, the DT classifier built on the optimal positive rate as well as the best resampling method. The resulting models are labelled as DT_SMOTE_50%_bagging and DT_SMOTE_50%_boosting, respectively. To make the comparison of all the established models easier, the performance of DT_SMOTE_50%_bagging and DT_SMOTE_50%_boosting is also illustrated in Figure 8. It is observed that comparing with individual DT, while Bagging on DT improves AUC, it hurts recall and F1 score, quite surprisingly. Compared to the base classifier DT, Boosting is beneficial when considering AUC, recall, and F1 score. Moreover, Boosting on DT wins Bagging by a huge margin in recall and F1 score, while Bagging outperforms Boosting by a margin in terms of AUC. In the stage 5 illustrated by Figure 3, by combining all the aforementioned discussions, we conclude that Boosting on DT by using the resampled training set with 50% positives via SMOTE method (i.e., DT_SMOTE_50%_boosting) is the optimal risk model in our study and it is labelled with a red star in Figure 8. It can achieve AUC, recall, and F1 score valued 0.8633, 0.9260, and 0.8907, respectively.

Figure 9 shows the variable importance on business risk by using Gini reduction in the DT_SMOTE_50%_boosting model. Higher score denotes more importance of the variable in classifying the business risk. Results show that three variables including MonLstRptDatePlcRec (i.e., number of months since the last report date on public records), NocurLiensJud (i.e., number of current liens or judgment), and MostRecentLienorJud (i.e., most recent lien or judgment) have the largest weight comparing with other variables. On the contrary, the variable such as pctSasNFA12mon (i.e., percentage of satisfactory non-financial accounts in the last 12 months) shows a negligible effect in the risk classifications. By ranking the risk factors, we can provide a comprehensive explanation on the possible reasons of business risk occurrence.

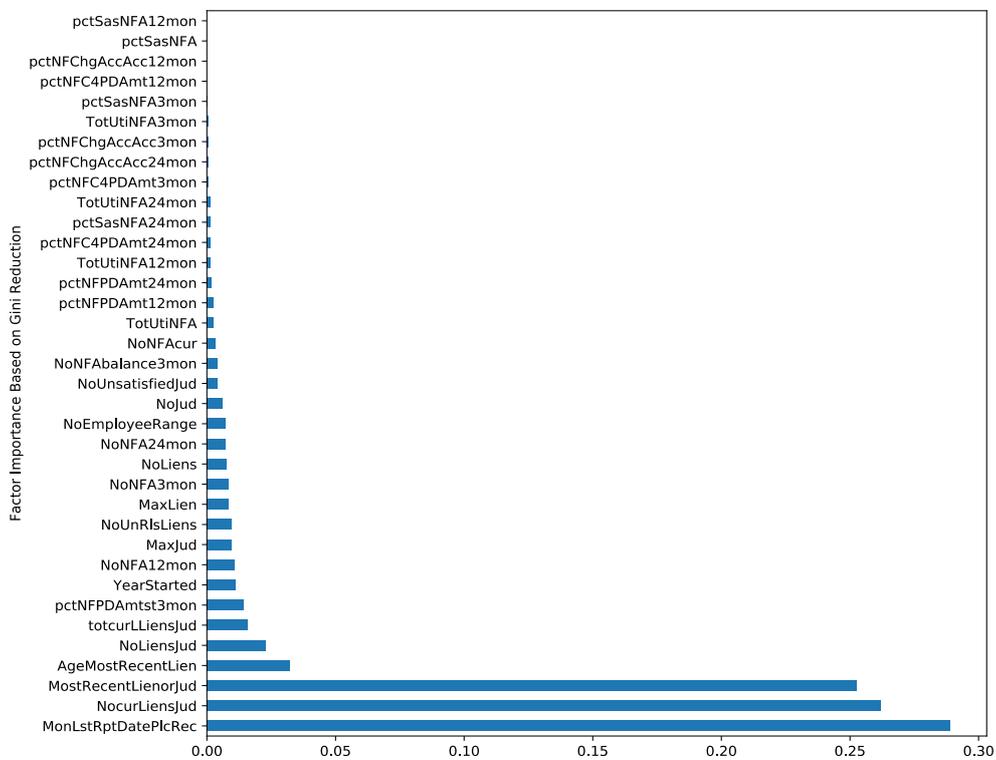

Figure 9. Identified business risk factors ranked by Gini reduction in the optimal model

## 5. CONCLUSION

In business risk classification tasks, the data used is often imbalanced and it arises many problems. In this study, we investigate how we can get a good risk model by simultaneously using a series of algorithms, including using proper model evaluation metrics, resampling along with different positive ratios, appropriate cross-validation, classifier regularization and ensembling. The simultaneous utilization of the above-mentioned algorithms is a major strength of this study, as there are few previous studies considering these techniques jointly.

In our experiment, AUC, rather than the widely-used accuracy, is selected as the primary measure when comparing different models. Two undersampling methods (RUS and CCUS) and two oversampling methods (ROS and SMOTE) are applied to resample the training data to reach a series of positive ratios ranging from 10% to 90% using a step of 10%. Three classifiers with strong interpretations including LR, L1LR, and DT are implemented on the resampled data as well as on the original data. Their performance is compared through 10-fold cross validation. Two model ensembling techniques, including Bagging and Boosting, are used on DT to further improve the DT performance. Finally, important features related to business risk are identified by using the optimal model.

Compared to the models built on the data without resampling, we found the increase in AUC, recall, and F1 score in all the three classifiers including LR, L1LR, and DT after resampling the training set to an appropriate positive percentage. SMOTE is shown to be the best resampling method for LR, L1LR, and DT across each percentage of positives in the training set. By using SMOTE, LR, L1LR, and DT can reach the highest AUC by using the resampled data with 20%, 20%, and 50% positives, respectively. On the other hand, among the four resampling methods, CCUS results in the lowest AUC in LR and DT when percentage of positives is between 20% and 80%. RUS and CCUS produce obviously different AUC on LR and DT. Although RUS and CCUS produce similar performance in L1LR, CCUS is less preferred because of its long processing time. Surprisingly, L1LR did not outperform LR in our result, since they produce very similar AUC under the same resampling procedure. By comparing with the base DT classifier, Boosting on DT is beneficial in terms of AUC, recall, and F1 score while Bagging on DT only improves AUC. The unexpected outcome is that Bagging on DT even slightly decrease recall and F1 score comparing to individual DT. The optimal candidate model for risk modeling is the Boosting model based on DT by using the resampled data with 50% positives via SMOTE. It can achieve the AUC, recall, and F1 score valued 0.8633, 0.9260, and 0.8907, respectively. MonLstRptDatePlcRec is shown to be the most important feature in classifying business risk while pctSasNFA12mon has little predictive power.

There is no general answer of the best resampling methods along with the resampled positive percentage in building risk models, as the answer is surely case and data dependent. However, the experimental design used in this study can serve as a reference for future studies in developing and improving risk models on imbalanced data. Moreover, the critical variables for the business risk classification provided by the Boosting DT model in our study could guide financial institutions in approving loan applications.

International Journal of Managing Information Technology (IJMIT) Vol.11, No.1, February 2019

**AUTHORS**

Yan Wang is a Ph.D. candidate in Analytics and Data Science at Kennesaw State University. Her research interest contains algorithms and applications of data mining and machine learning techniques in financial areas. She has been a summer Data Scientist intern at Ernst & Young and focuses on the fraud detections using machine learning techniques. Her current research is about exploring new algorithms/models that integrates new machine learning tools into traditional statistical methods, which aims at helping financial institutions make better strategies. Yan received her M.S. in Statistics from university of Georgia.

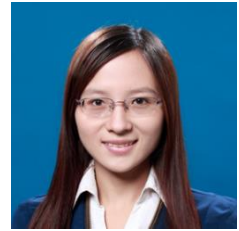

Dr. Xuelei Sherry Ni is currently a Professor of Statistics and Interim Chair of Department of Statistics and Analytical Sciences at Kennesaw State University, where she has been teaching since 2006. She served as the program director for the Master of Science in Applied Statistics program from 2014 to 2018, when she focused on providing students an applied leaning experience using real-world problems. Her articles have appeared in the Annals of Statistics, the Journal of Statistical Planning and Inference and Statistica Sinica, among others. She is the also the author of several book chapters on modeling and forecasting. Dr. Ni received her M.S. and Ph.D. in Applied Statistics from Georgia Institute of Technology.

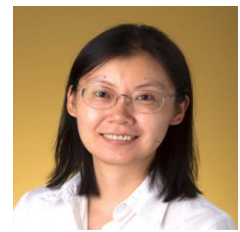